\documentclass[letterpaper, 10 pt, conference]{ieeeconf}  % Comment this line out if you need a4paper

\IEEEoverridecommandlockouts                              % This command is only needed if 
\usepackage{cite}
\usepackage{amsmath,amssymb,amsfonts}
\usepackage{algorithmic}
\usepackage{graphicx}
\usepackage{textcomp}
\usepackage{booktabs}
\usepackage[ruled,vlined]{algorithm2e}

\usepackage{mathtools}
\DeclarePairedDelimiterX{\norm}[1]{\lVert}{\rVert}{#1}

\usepackage{amsthm}
\usepackage[font=small]{caption}
\usepackage{subfigure}
\usepackage{graphics} % for pdf, bitmapped graphics files
\usepackage{epsfig} % for postscript graphics files
\usepackage{array}
\usepackage{url}
\usepackage{diagbox}
\usepackage{epstopdf}
\usepackage{paralist}
\AppendGraphicsExtensions{.tif}
\usepackage{enumerate}
\usepackage{import}
\usepackage{caption}
\DeclareMathOperator*{\argmin}{arg\,min}

\makeatletter
\renewcommand*\env@matrix[1][\arraystretch]{%
  \edef\arraystretch{#1}%
  \hskip -\arraycolsep
  \let\@ifnextchar\new@ifnextchar
  \array{*\c@MaxMatrixCols c}}
\makeatother

\title{\LARGE \bf
Learning-based Uncertainty-aware Navigation in 3D Off-Road Terrains}

\author{Hojin Lee, Junsung Kwon, and Cheolhyeon Kwon% <-this % stops a space
\thanks{This work was supported in part by National Research Foundation of Korea (NRF) funded by the Korea Government (MSIT) under Grant 2020R1C1C1007323, and in part by the 2022 Research Fund (1.220025.01) of UNIST (Ulsan National Institute of Science and Technology).}
\thanks{The authors are with the Department of Mechanical Engineering, Ulsan National Institute of Science and Technology, Ulsan, 44919 Repulic of Korea (e-mail:hojinlee@unist.ac.kr; kjs1145@unist.ac.kr; kwonc@unist.ac.kr).}%
}

\begin{document}

\maketitle
\thispagestyle{empty}
\pagestyle{empty}

%%%%%%%%%%%%%%%%%%%%%%%%%%%%%%%%%%%%%%%%%%%%%%%%%%%%%%%%%%%%%%%%%%%%%%%%%%%%%%%%
\begin{abstract}
This paper presents a safe, efficient, and agile ground vehicle navigation algorithm for 3D off-road terrain environments. Off-road navigation is subject to uncertain vehicle-terrain interactions caused by different terrain conditions on top of 3D terrain topology. The existing works are limited to adopt overly simplified vehicle-terrain models. The proposed algorithm learns the terrain-induced uncertainties from driving data and encodes the learned uncertainty distribution into the traversability cost for path evaluation. The navigation path is then designed to optimize the uncertainty-aware traversability cost, resulting in a safe and agile vehicle maneuver. Assuring real-time execution, the algorithm is further implemented within parallel computation architecture running on Graphics Processing Units (GPU).
\end{abstract}

%%%%%%%%%%%%%%%%%%%%%%%%%%%%%%%%%%%%%%%%%%%%%%%%%%%%%%%%%%%%%%%%%%%%%%%%%%%%%%%%
\section{Introduction}
Autonomous navigation of ground vehicles in an off-road environment has been of vital importance in many applications, such as military operations, agriculture, and planet exploration \cite{papadakis2013terrain,mousazadeh2013technical,kantor2003distributed}. Compared to paved road cases \cite{betz2022autonomous,gonzalez2015review}, off-road terrains impose an additional challenge to autonomous navigation. This is mainly attributed to the complicated vehicle-terrain interactions, characterized by different factors such as terrain topology, terrain type, etc \cite{papadakis2013terrain}. These factors are further subject to various uncertainties, making it difficult to assess the vehicle maneuver in reality. To safely and agilely guide the vehicle under such uncertain vehicle-terrain interactions, this paper mainly investigates the following two problems.
\begin{itemize}
\item 
Planning the feasible path of the off-road vehicle while addressing nonlinear vehicle dynamics over 3D terrain topology.
\item
Accounting for the uncertainties of vehicle maneuver in different terrain conditions, thereby generating a safe yet less-conservative path.
\end{itemize}

\begin{figure}[ht]
\centering
\includegraphics[width=0.45\textwidth]{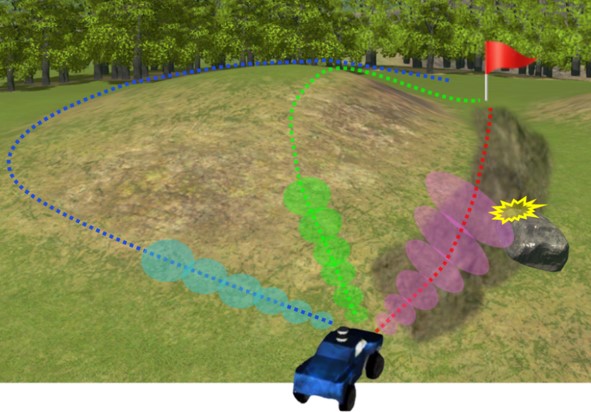}
\caption{Example paths along varying terrains. The shortest path (red line) encounters mud areas with higher motion uncertainty (red-ellipsoidal area), which has a higher chance of slipping and colliding with an obstacle. The path detouring the hill (blue line) is completely getting around the mud and slop at the cost of a long path. The proposed algorithm can appropriately compromise both paths and generate an agile yet safe path (green line).
}\label{fig:mainfigure}
\end{figure}

It is worth noting that the existing work focused on either one problem while treating another in an overly simplified sense. For example, some approaches generate a safe path considering 3D terrain topology and 6 DOF vehicle dynamics while ignoring the uncertainty caused by the types of terrains \cite{yu2021nonlinear,shen2021three,krusi2017driving,hu2021integrated}. On the other hand, other approaches assess the path by different terrain conditions(e.g., roughness, slip coefficients) and apply a simple kinematic model of the vehicle with the assumption of driving on flat terrain \cite{mizuno2020new,kahn2021badgr,cai2022risk,gasparino2022wayfast,manderson2020learning,gregory2021improving}. Such approaches may be acceptable to generate safe paths in a simple environment. In practice, however, a lack of simultaneous consideration of the above two problems may result in a physically infeasible or too conservative path (See Fig. \ref{fig:mainfigure}).

This paper proposes a navigation algorithm that can agilely guide a vehicle to the goal while accommodating the uncertainty in the 3D terrain with different terrain conditions (e.g., slop, friction coefficient, roughness). First, raw sensor measurements are processed to construct a geometric traversability cost map, and the terrain type map encodes semantic features (e.g., grass, mud, asphalt, etc.). Then, the vehicle-terrain interactions for individual terrain types are respectively learned through Gaussian Process (GP) regression models. By virtue of the GP models, the terrain-induced uncertainties can be expressed by probability distributions. These distributions are used to predict the actual path distributions of the vehicle when the vehicle follows the candidate path, resulting in \emph{predictive path distributions} \cite{capone2020anticipating}. Next, the best path is found by evaluating the cost metrics associated with traversability, rollover risk, and distance to the goal. Note that the cost evaluation comprehends the terrain-induced uncertainty distribution via the Gaussian kernel smoothing technique \cite{nixon2019feature}.

The proposed algorithm involves high-dimensional nonlinear dynamics in multiple GP models, making it computationally expensive. To run the algorithm through an onboard computing unit of the vehicle, we resort to sampling-based methods. For instance, a finite number of candidate paths are generated based on the control input sampler. And the predictive path distribution for each path candidate is computed based on sample propagation. Lastly, the sampling-based nonlinear model predictive control (MPC) is applied to follow the resulting path. Furthermore, we leverage the recent advances in GPU parallel computing, which can effectively implement sampling-based algorithms for real-time operation.

\section{Related Works}

\subsection{Vehicle Navigation over 3D Terrains}
Early works have addressed the navigation on 3D terrains based on the 2D navigation methods while simplifying the original 3D problem. In \cite{wang2019safe}, a multi-layer 2D map extracted from the 3D OctoMap has been utilized. In \cite{jian2022putn}, a path sampling algorithm based on the locally fitted plane has been proposed to generate a global path. They applied GP model to interpolate the local paths computed from each locally fitted plane to make a denser path. In \cite{rastgoftar2018data,krusi2017driving}, safe navigation areas have been classified based on the labeled 3D point cloud map given by the Lidar sensor. However, such a binary division cannot preclude dynamically infeasible paths or collisions due to the complex vehicle interaction with 3D terrain topology. A more general global path planning algorithm has been proposed in \cite{hu2021integrated}. This method utilizes a potential field function and leverages the Voronoi diagram to improve the computational efficiency while considering the 3D kinematic vehicle model and the terrain topological constraints. 

From the perspective of optimal control and local planning, MPC-based approaches have been developed for off-road environments but are limited to driving on flat terrains \cite{liu2018nonlinear,febbo2017moving,liu2017combined}. In \cite{yu2021nonlinear}, the author formulated a nonlinear MPC problem for optimal maneuvers on 3D terrains. In this approach, the terrain topology is assumed to be twice continuously differentiable, which might be invalid in practice. Very recently, a learning-based method has indirectly considered the effects of 3D terrains by utilizing the elevation map, and 3D vehicle poses as training features \cite{weerakoon2022terp}. To sum up, all the aforementioned methods lack the comprehension of the non-linearity of the 3D vehicle dynamics and/or the effect of different terrain conditions.

\subsection{Vehicle Navigation in View of Traversability}
A plethora of research has been carried out to address the effect of vehicle-terrain interactions by the name of terrain traversability. The main idea is to evaluate the traversability of the terrain areas and safely navigate only through the traversable areas. The methods for determining the traversable area can be divided into three main categories; i) \emph{geometry-based}; ii) \emph{semantic-based}; and iii) \emph{learning-based}. 

First, the geometry-based methods examine the traversability according to the 3D geometry of the surface, i.e., slope, step height, and roughness. The elevation map, created by Lidar point clouds \cite{wermelinger2016navigation,fankhauser2018probabilistic}, has been used to compute the geometrical features of the 3D surface \cite{leung2022hybrid,zhao2019semantic}. Using geometry-based traversability cost, local and global navigation algorithms have been proposed in \cite{thoresen2021path,overbye2020fast,jian2022putn}. However, such methods disregard the vehicle-terrain interactions derived from the terrain conditions, such as friction coefficients. 

Meanwhile, the semantic-based methods typically analyze the types of terrain segmented from the camera image \cite{dabbiru2021traversability,nardi2019actively,maturana2018real,leung2022hybrid}. For example, in \cite{guantns,maturana2018real,leung2022hybrid}, the traversability is assigned manually based on the terrain types, i.e., low traversability in the mud area and high traversability in the grass area. The limitation of such methods comes from the fact that heuristic user-interventions are inevitable when designing the traversability cost. 

Recent learning-based methods can avoid such heuristics as they exploit the collected data to learn the traversability cost instead of manual cost assignment \cite{nardi2019actively,waibel2022rough,kahn2021badgr,cai2022risk,gasparino2022wayfast}. In \cite{nardi2019actively,waibel2022rough,manderson2020learning}, the roughness of the terrains is learned based on the sensor data, and they plan a smoother path for the vehicle. In \cite{cai2022risk}, a neural network represents traversability as the achievable maximum speed distribution of vehicles on different terrains. Moreover, end-to-end learning approaches have been proposed to avoid geometric obstacles as well as untraversable terrains by analyzing the executed commands and the realized trajectories \cite{kahn2021badgr,gasparino2022wayfast}. Despite rich literature, the developed learning-based methods do not explicate 3D terrain topology, whose impact is too costly to learn from the collected data only.

\section{Problem Formulation}
This section presents the notations, and dynamical vehicle model on 3D terrain topology, followed by the GP model for analyzing the vehicle-terrain interactions, which is the key ingredient of our proposed algorithm.  

\subsection{Notations and Definitions}
Let us denote a world reference frame as $\{O^{\mathcal{W}},X^{\mathcal{W}},Y^{\mathcal{W}},Z^{\mathcal{W}}\}$ with superscript $\mathcal{W}$, and a vehicle reference frame as
$\{O^\mathcal{B},X^\mathcal{B},Y^\mathcal{B},Z^\mathcal{B}\}$ with superscript $\mathcal{B}$. And the vehicle orientation with respect to the world reference frame is defined by a set of Euler angles, roll$(\phi)$, pitch$(\theta)$, and yaw$(\psi)$. It is assumed that the roll and pitch of the vehicle can be determined by the surface normal vector and the yaw of the vehicle. The rotation matrix $R^{\mathcal{W}}_{\mathcal{B}}$ defines the transformation from vehicle to world reference frame as follows.
\begin{equation*}
    R^{\mathcal{W}}_{\mathcal{B}} = \begin{bmatrix}
c\psi c\theta & c\psi s\theta s\phi -s\psi c\phi & c\psi s \theta c \phi  + s \psi s \phi \\
s\psi c \theta & s\psi s \theta s \phi +c \psi c \phi & s \psi s \theta c \phi -c \psi s \phi \\ 
-s \theta  & c \theta s \phi  & c \theta c \phi 
\end{bmatrix}
\end{equation*}
where $c\ast$ and $s\ast$ are shorthand notations for $cos(\ast)$ and $sin(\ast)$, respectively. Then the gravitational force acting on the vehicle to its center of mass can represented as $F^{\mathcal{B}} = (R^{\mathcal{W}}_{\mathcal{B}})^{-1}F^{\mathcal{W}}$ where $F^{\mathcal{W}} = [0,0,-mg]^{\mathrm{T}}$ is the weight force in world reference frame. The velocities of vehicle in the world frame are denoted by $v_{x}^{\mathcal{W}}$, $v_{y}^{\mathcal{W}}$, $v_{z}^{\mathcal{W}}$. And the longitudinal, lateral, and vertical velocity components in the body-fixed frame are denoted by $v_{x}^{\mathcal{B}}$, $v_{y}^{\mathcal{B}}$, and $v_{z}^{\mathcal{B}}$. Angular velocities in the world reference frame can be computed as 
\begin{equation*}
    \begin{bmatrix}[1.2]
    \dot{\phi} \\ 
    \dot{\theta} \\ 
    \dot{\psi} 
    \end{bmatrix} 
    = 
    \begin{bmatrix}[1.2]
  1 & \frac{s\phi*s\theta}{c\theta} & \frac{c\phi*s\theta}{c\theta} \\ 
       0 & c\phi & -s\phi \\ 
       0 & \frac{s\phi}{c\theta} & \frac{c\phi}{c\theta}
    \end{bmatrix}
      \begin{bmatrix}[1.2]
  \omega_{x}^{\mathcal{B}} \\
  \omega_{y}^{\mathcal{B}} \\  
  \omega_{z}^{\mathcal{B}}
    \end{bmatrix}
\end{equation*}
where $\omega_{x}^{\mathcal{B}}$, $\omega_{y}^{\mathcal{B}}$, $\omega_{z}^{\mathcal{B}}$ are the angular velocity in the body-fixed frame. Without loss of generality, the angular velocities $\omega_{x}^{\mathcal{B}}$, and $\omega_{y}^{\mathcal{B}}$ are approximated to be zero \cite{yu2021nonlinear}. $L_{f}$, and $L_{r}$ are the distance to the front and rear axles from the center of gravity (c.g.) of vehicle. $L_{w}$ is the track width and $h$ is the height of the c.g. of the vehicle
measured from the ground. And $h_{R}$ is the height of the c.g. from the roll center. $m$ is the vehicle mass and $I_{zz}$ is the yaw moment of inertia. $0_{a}\in\mathbb{R}^{a\times a}$ and $I_{a}\in\mathbb{R}^{a\times a}$ are zero and identity matrix.

\subsection{Vehicle Dynamics over 3D Terrains}\label{pf:vehicle_dynamics}
In this paper, the extended dynamic bicycle model is considered \cite{yu2021nonlinear}. Correspondingly, the vehicle state is defined as $\xi := \begin{bmatrix} x^{\mathcal{W}} \ \ y^{\mathcal{W}} \ \ \psi \ \ v_{x}^{\mathcal{B}} \ \ v_{y}^{\mathcal{B}} \ \  \omega_{z}^{\mathcal{B}} 
    \end{bmatrix}^{\mathrm{T}} \in\mathbb{R}^{6}$, where $x^{\mathcal{W}}$, $y^{\mathcal{W}}$ represents the global position of the vehicle along $X^{\mathcal{W}}$, and $Y^{\mathcal{W}}$ axis, respectively. And the control inputs are denoted as $\zeta :=\begin{bmatrix} \delta \ \  a_{x}^{\mathcal{B}}
    \end{bmatrix}^{\mathrm{T}}$
where $\delta$ is the steering angle and $a_{x}^{\mathcal{B}}$ is the longitudinal acceleration. Then, the vehicle dynamics is described by 
\begin{equation}\label{eq:vehicle_dynamics}
\dot{\xi} = f(\xi,\zeta) = 
\begin{bmatrix}[1.4]
  v_{x}^{\mathcal{W}} \\ 
  v_{y}^{\mathcal{W}} \\
  \frac{c\phi}{c\theta}\omega_{z}^{\mathcal{B}}\\
  a_{x}^{\mathcal{B}} \\ 
  \frac{F_{yf}+F_{yr}-F_{G}}{m}-v_{x}^{\mathcal{B}}\omega_{z}^{\mathcal{\mathcal{B}}}\\
  \frac{F_{yf}L_{f}cos(\delta) - L_{r}F_{yr}}{I_{zz}}
\end{bmatrix}    
\end{equation}
where $F_{G}= -mg\times c\theta s\phi$ is the component of vehicle gravity force in the lateral direction. $F_{yf}$, and $F_{yr}$ are the lateral tire forces on the front and rear tires calculated using the linear tire model as follows.
\begin{equation*}
        F_{yf}  = C_{\alpha f}F_{zf}\alpha_{f},\  \         F_{yr}  = C_{\alpha r}F_{zr}\alpha_{r}
\end{equation*}
where $C_{\alpha f}$, and $C_{\alpha r}$ are the front and rear cornering stiffness and $\alpha_{f}$, $\alpha_{r}$ are the front and rear tire slip angles. And the vertical loads can be drawn in the body-fixed frame as follows \cite{doumiati2009lateral}.
\begin{equation*}
\begin{split}
    F_{zf} &=
    \frac{L_{r}mg\times c\theta c\phi+hma_{x}^{\mathcal{B}}+hmg\times s\theta}{L_{r}+L_{f}}\\ 
    F_{zr} &= 
    \frac{L_{f}mg\times c\theta c\phi-hma_{x}^{\mathcal{B}}-hmg\times s\theta}{L_{r}+L_{f}}
\end{split}
\end{equation*}

\subsection{Gaussian Process for Vehicle-Terrain Interaction}\label{ssc:gpmodel}
In this section, GP model is presented to infer the vehicle-terrain interactions. We depict the vehicle-terrain interaction as the modeling errors of vehicle velocities, i.e., $(v_{x}^{\mathcal{B}}, v_{y}^{\mathcal{B}}, \omega_{z}^{\mathcal{B}})$. A similar approach has been delivered in \cite{gasparino2022wayfast}, which estimates the traversable cost in terms of modeling errors. However, they only utilized the kinematic model assuming the flat surface. Let us denote the velocity error between true vehicle and the nominal system model in \eqref{eq:vehicle_dynamics} at the discrete time step $k$ as follows.
\begin{equation*}\begin{split}
    e_{vx}(k) &=v_{x}^{\mathcal{B}}(k)-\hat{v}_{x}^{\mathcal{B}}(k) \\
    e_{vy}(k)&= v_{y}^{\mathcal{B}}(k)-\hat{v}_{y}^{\mathcal{B}}(k) \\
    e_{\omega}(k) &= \omega_{z}^{\mathcal{B}}(k)-\hat{\omega}_{z}^{\mathcal{B}}(k)
\end{split}
\end{equation*}
where $\hat{v}_{x}^{\mathcal{B}}(k)$, $\hat{v}_{y}^{\mathcal{B}}(k)$, and $\hat{\omega}_{z}^{\mathcal{B}}(k)$ are the predicted states propagated by \eqref{eq:vehicle_dynamics} with $\xi(k-1)$, and $\zeta(k-1)$. Then the GP model is trained to represent the following vector-valued function.
\begin{equation*}
    g(x_{gp}(k)) =  y(k) = [e_{vx}(k+1) \ \ e_{vy}(k+1) \ \  e_{\omega}(k+1)]^{\mathrm{T}}
\end{equation*}
where $x_{gp} = [v_{x}^{\mathcal{B}} \ v_{Y}^{\mathcal{B}} \ \omega_{z}^{\mathcal{B}} \ \phi \ \theta  \ \delta \  a_{x}^{\mathcal{B}}]^{\mathrm{T}}$ and $y(k)$ are the model input and output measurement vectors, respectively. A GP for a function $g$ is set to have zero prior mean and a squared exponential kernel function $\kappa : \mathbb{R} \times \mathbb{R} \rightarrow \mathbb{R}$ as follows \cite{williams2006gaussian}. 
\begin{equation*}
    \kappa(x_{gp},x_{gp}^{'}) = \sigma^{2}
    exp\left(- 
    \frac{(x_{gp}-x_{gp}^{'})^{\mathrm{T}}L^{-2}(x_{gp}-x_{gp}^{'})}
    {2}\right)
\end{equation*}
where $\sigma$, and $L$ are the hyperparameters trained by maximizing the log marginal likelihood.
\begin{figure*}[ht]
\centering
\includegraphics[width=0.99\textwidth]{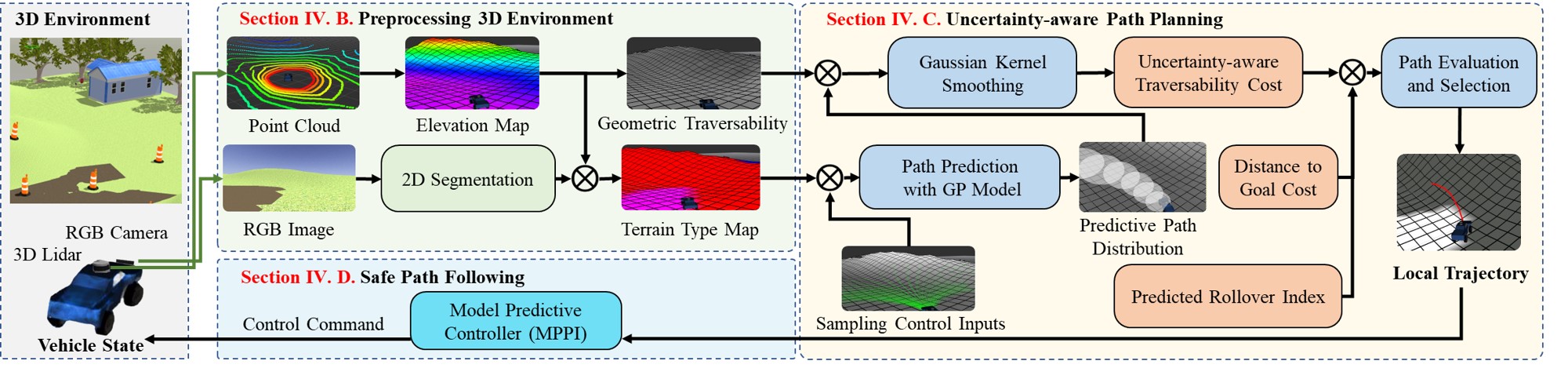}
\caption{Overview of the learning-based uncertainty-aware navigation in 3D terrain}\label{fig:algorithm_overview}
\end{figure*}
Based on the data of measurements $y$ corresponding to the inputs $X$, the predictions of GP at $x_{gp}^{\ast}$ are represented by the predictive mean $\mu_{\ast}$ and variance $\sigma_{\ast}^{2}$ defined as: 
\begin{equation}\label{eq:gpmodel}
    \begin{split}
    \mu_{\ast} &= K(x_{gp}^{\ast},X)K_{\ast}^{-1}y\\
    \sigma_{\ast}^{2} &= K(x_{gp}^{\ast},x_{gp}^{\ast}) -K(x_{gp}^{\ast},X)K_{\ast}^{-1}K(X,x_{gp}^{\ast})
    \end{split}
\end{equation}
where $K_{\ast} = K(X,X)$, and $K(\cdot,\cdot)$ are matrices constructed using the kernel $\kappa(\cdot,\cdot)$ evaluated at the test and training data, $x_{gp}^{\ast}$, and X. Our proposed algorithm utilizes multiple number of GP models such that $g_{j}, j\in\{1,\cdots,J\}$ represents the vehicle-terrain interaction on $J$ different types of terrains, e.g., grass, rock, and mud etc.

\section{Algorithm Development}
This section describes sub-modules of the proposed learning-based uncertainty-aware navigation algorithm in detail. 
\subsection{Algorithm Overview}
The algorithm is decomposed into three main modules; 1) preprocessing 3D environment; 2) uncertainty-aware path planning; and 3) safe path following. The preprocessing step outputs the geometric traversability cost and terrain type map based on the RGB camera images and 3D Lidar sensor point clouds. The path planning module generates the predictive path distributions from the sampled control inputs using the learned vehicle-terrain interaction model. Then the generated predictive path distributions are evaluated to select the best path based on the uncertainty-aware traversability cost, the risk of rollover, and distance to the goal. The last module solves the optimal control problem to follow the resulting path using a 6 DOF vehicle model in \eqref{eq:vehicle_dynamics}. The overview of the algorithm is illustrated in Fig. \ref{fig:algorithm_overview}.

\subsection{Preprocessing 3D Environment}\label{ssc:preprocessing}
This module processes raw sensor measurements and generates global terrain grid maps. First, the elevation map developed by \cite{fankhauser2014robot} is created using 3D point clouds from the LIDAR sensor. The created elevation map saves the height information in each grid. Then, the geometric traversability cost is computed for each grid as follows. 
\begin{equation*}    
    T_{geo} = w_{1}T_{s} + w_{2}T_{r} + w_{3}T_{h}
\end{equation*}
where $w_{1},w_{2}$, and $w_{3}$ are scale factors summing up to $1$. And $T_{s}$, $T_{r}$, and $T_{h}$ are the slop, roughness, and step height costs computed similar to \cite{wermelinger2016navigation}. We also set a robot-specific maximum threshold for traversability cost, so the total cost is bounded below the threshold. 

Furthermore, terrain types are extracted from the RGB camera image with the help of the semantic segmentation module, i.e., \cite{romera2017erfnet,takikawa2019gated}. Each pixel is classified into different types of terrains from a predefined set of terrain categories. Then, the 2D pixels are projected to a 3D world coordinate frame \cite{leung2022hybrid}. Consequently, 
each grid in a elevation map can be labeled with types of terrain to form a terrain type map.

\subsection{Uncertainty-aware Path Planning}
\begin{figure}[ht]
\centering
\includegraphics[width=0.49\textwidth]{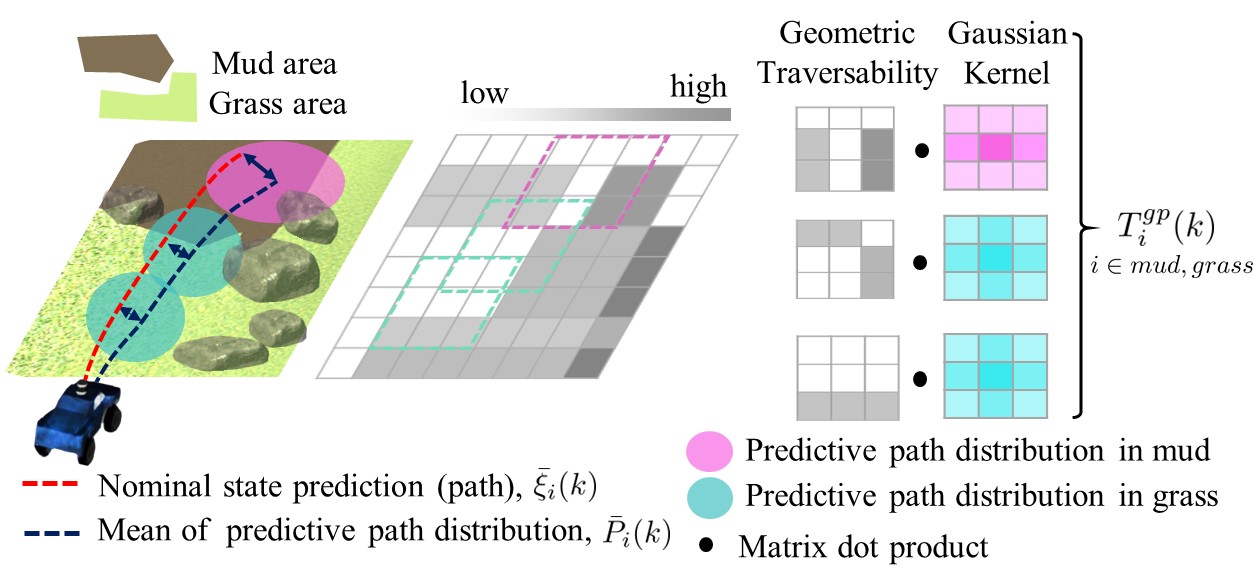}
\caption{Illustration of traversability cost computation for the given predictive distribution of vehicle poses at each step.
}\label{fig:trav_cost}
\end{figure}

In this module, the predictive path distributions corresponding to the individual control input samples are computed. Given the initial vehicle state $\xi(0)$, we sample $I$ number of control inputs as $\zeta_{i}, i\in\{1,\cdots,I\}$. And we sample a collection of nominal state predictions (i.e., nominal path) with $N$ time step prediction horizons, each matched to each control input.
\begin{equation}\label{eq:nominalpath}
    \bar{\xi}_{i}(k),\  i\in\{1,\cdots,I\} ,\   k\in\{0,\cdots, N\}
\end{equation}
where each state is propagated by the nonlinear vehicle dynamics in  \eqref{eq:vehicle_dynamics}.
\begin{equation*}
    \bar{\xi}_{i}(k+1) = f(\bar{\xi}_{i}(k),\zeta_{i})
\end{equation*}
It is worth noting that one can analyze the traversability of the sampled nominal state predictions, $\bar{\xi}_{i}(k)$, by evaluating the costs from the created traversability maps that underlie the nominal path of the state predictions. However, this idea is prone to fail due to the uncertain vehicle-terrain interactions in the real world. Let us consider the vehicle that travels along the mud area near the obstacle, as shown in Fig. \ref{fig:trav_cost}. There will be a higher chance that the vehicle collides with the obstacle as the nominal path does not align with the actual path due to the slip or the deficient performance of the following control.

To overcome such limitations, our proposed algorithm learns the discrepancy between the nominal and actual path, and accommodates it in the planning phase. We first draw the predictive path distributions learned from the driving data. The GP model \eqref{eq:gpmodel} is employed to generate $M$ samples of actual state prediction (i.e., predictive path) for each control input.
\begin{equation*}
    \xi_{i}^{(m)}(k),\  i\in\{1,\cdots,I\} ,\   k\in\{0,\cdots, N\},\  m\in\{1,\cdots,M\}
\end{equation*}
where superscript $m$ represents the predictive path sample index. This time, the state update can be derived by applying \eqref{eq:vehicle_dynamics} and adding a sample output from the posterior GP distribution, represented by the following stochastic dynamical model.
\begin{equation*}
\xi_{i}^{(m)}(k+1) = f(\xi_{i}^{(m)}(k),\zeta_{i}) + B\
g_{j}(F\xi_{i}^{(m)}(k),\psi,\theta,  \zeta_{i})
\end{equation*}
where $B = [0_{3};I_{3}]$ and $F = [0_{3}\ I_{3}]$. GP model, $g_{j}$, is selected based on the terrain type retrieved from the terrain type map at the vehicle position. Furthermore, we denote the position of each sample as $P_{i}^{(m)}(k)=[I_{2}\ 0_{4}]\times \xi_{i}^{(m)}(k)\in\mathbb{R}^{2}$. Correspondingly, the mean and variance of predictive path distribution of vehicle can be computed as follows \cite{brudigam2021gaussian,capone2020anticipating}.
\begin{equation}\label{eq:trajdistribution}
    \begin{split}
        \bar{P}_{i}(k) &= \frac{ \sum_{m=1}^{M}P_{i}^{(m)}(k)}{M}, \ k\in\{1,\cdots,N\}\\
        \Sigma_{i}(k) & = \frac{\sum_{m=1}^{M}\left(
        P_{i}^{(m)}(k) - \bar{P}_{i}(k)
        \right)^{\mathrm{T}}
        \left(
        P_{i}^{(m)}(k) - \bar{P}_{i}(k)
        \right)}{M-1}
    \end{split}
\end{equation}
To compute the traversability cost over the predictive path distribution, we borrow the idea from 2D Gaussian kernel smoothing \cite{nixon2019feature}. We construct a 2D Gaussian kernel matrix  $\mathcal{K}_{i}(k)\in\mathbb{R}^{s\times s}$ for each position using $\bar{P}_{i}(k)$ and  $\Sigma_{i}(k), \ k \in\{1,\cdots,N\}$ where $s$ is the predefined kernel size. And we denote $T_{geo}(\bar{P}_{i}(k)) \in \mathbb{R}^{s\times s} $ as the local submap of geometrical traversability cost map centered at $\bar{P}_{i}(k)$ (See Fig. \ref{fig:trav_cost}). Then, one can define the predictive geometric traversability cost for the predictive path distribution with the mean $\bar{P}_{i}(k)$, and variance $\Sigma_{i}(k)$ as follows.
\begin{equation*}
    T^{gp}_{i}(k) := 1_{s}^{\mathrm{T}}\times\left(\mathcal{K}_{i}(k) \cdot T_{geo}(\bar{P}_{i}(k))\right)\times1_{s}
\end{equation*}
where $1_{s}$ is the column vector whose entries are all 1's. To prevent the actual path from excessively deviating from the nominal path, we further penalize the nominal path \eqref{eq:nominalpath} for the errors of the corresponding predictive path distribution \eqref{eq:trajdistribution}. Specifically, the following predictive uncertainty cost is defined by Mahalanobis distance \cite{mahalanobis1936generalized}.
\begin{equation}\label{eq:predictiveuncertaintycost}
    e_{i}(k) := \sqrt{\left(B_{e}\bar{\xi}_{i}(k) - \bar{P}_{i}(k)\right)^{\mathrm{T}} \Sigma_{i}^{-1}(k)\left(B_{e}\bar{\xi}_{i}(k) - \bar{P}_{i}(k)\right)}
\end{equation}
where $B_{e} = [I_{2} \ 0_{4}]\in\mathbb{R}^{2\times 6}$. Again, the cost in \eqref{eq:predictiveuncertaintycost} is influenced by various factors, including vehicle-terrain interaction, the topology of 3D terrain, and even the capability of the following controller. To sum up, the uncertainty-aware traversability cost for each control input sample can be computed as follows. 
\begin{equation*}
    T_{i} = \sum_{k=1}^{N}\left(w_{gp}T_{i}^{gp}(k) + w_{e}e_{i}(k) \right), \ i\in\{1,\cdots,I\}
\end{equation*}
where $w_{gp}$ and $w_{e}$ are the weighting parameters. 

Apart from the uncertainty-aware traversability cost, the rollover risk is accounted for planning phase to assure safety during the agile motion of the vehicle. The Rollover index is used to express the vehicle status whether the vehicle reaches the critical rollover point \cite{qian2020rollover,lefevre2014survey}. Given the vehicle states, the corresponding rollover index is computed by
\begin{equation}\label{eq:rollover}
    R_{roll}(\xi) = \frac{2[(ma_{y}^{\mathcal{B}}+F_{G})h_{R} - mgh_{R}\times s\psi)
    ]}{L_{w}(F_{zf}+F_{zr})}  
\end{equation}
where
$a_{y}^{\mathcal{B}} = \frac{C_{\alpha f}}{m}
(\delta-\frac{v_{y}^{\mathcal{B}}+L_{f}\omega_{z}^{\mathcal{B}}}{v_{x}^{\mathcal{B}}})
+ \frac{C_{\alpha r}}{m}
\frac{L_{r}\omega_{z}^{\mathcal{B}}-v_{y}^{\mathcal{B}}}{v_{x}^{\mathcal{B}}}
$ is the lateral acceleration in vehicle reference frame \cite{rajamani2011vehicle}. If any state predictions given the control inputs induce a rollover index above the safety threshold, i.e., $R_{roll}(\bar{\xi}_{i}^{(m)}(k)) > R_{thres}$, we exclude the corresponding path to ensure safety.

The final step of the planning module is selecting the best local path to guide a vehicle to the goal with the least traversability cost. For each nominal path generated by the $i^{th}$ sampled control input, one can compute the closest Euclidean distance from the sampled nominal path to the goal as $T^{dist}_{i}$. Then, the best local path is selected as follows.
\begin{equation*}
    \bar{\xi}_{i^{\ast}}(k), \ k\in\{0,\cdots,N\}, \  i^{\ast} = \argmin_{\forall i \in  \Omega_{Safe}} \left(T_{i} + w_{dist}\times T^{dist}_{i}\right)
\end{equation*}
where $w_{dist}$ is the weighting parameter. And $\Omega_{Safe}$ is the index set of control inputs which do not violate rollover condition \eqref{eq:rollover} along the path.
\begin{equation*}
\begin{split}
\Omega_{Safe} := \{i\in&\{1,\cdots,I\}|R_{roll}(\bar{\xi}_{i}^{(m)}(k)) \leq R_{thres},\\ & k\in\{1,\cdots,N\}, m\in\{1,\cdots,M\}\}
\end{split}
\end{equation*}
 
\subsection{Safe Path Following}
Once the best path is selected, the path following algorithm is implemented based on a nonlinear MPC scheme. Given the current vehicle state and the local path $\bar{\xi}_{i^{\ast}}(k), \ k\in\{0,\cdots,N\}$, a finite horizon optimization problem is formulated to obtain a sequence of control input that minimizes the cost function as follows.
\begin{equation}\label{eq:mppiCost}
\begin{split}
    \min\limits_{\xi(k),\zeta(k)} \sum_{k=0}^{\bar{N}-1}
    \big\{ || \xi(k)&- \bar{\xi}_{i^{\ast}}(k)||^{2}_{\mathcal{Q}} + ||\zeta(k)||^{2}_{\mathcal{R}} 
    \big\}\\
    &+ || \xi(\bar{N})- \bar{\xi}_{i^{\ast}}(\bar{N})||^{2}_{\mathcal{Q}}    
\end{split}
\end{equation}
where $\mathcal{Q}$ and $\mathcal{R}$ are the cost weight matrices. And $\bar{N}$ is the prediction horizon for the MPC. The solution for the optimization problem in \eqref{eq:mppiCost} is denoted as $\zeta^{\ast}(k), \ k\in\{0,\cdots,\bar{N}-1\}$ and the first sequence $\zeta^{\ast}(0)$ is applied to the vehicle. Since the dynamics are not twice continuously differentiable in the sharp elevation map, we utilize a sampling-based MPC method called model predictive path integral control (MPPI) \cite{williams2017model,williams2017information}. In short, this controller generates sample trajectories which are evaluated to find the minimum cost trajectory and the corresponding inputs. As sampling can be easily parallelized on the GPU, we can sample 5000 trajectories at about 10Hz in real-time.

\section{Experiments}

\subsection{Simulation Setup}
\begin{figure}[ht]
\centering
\includegraphics[width=0.38\textwidth]{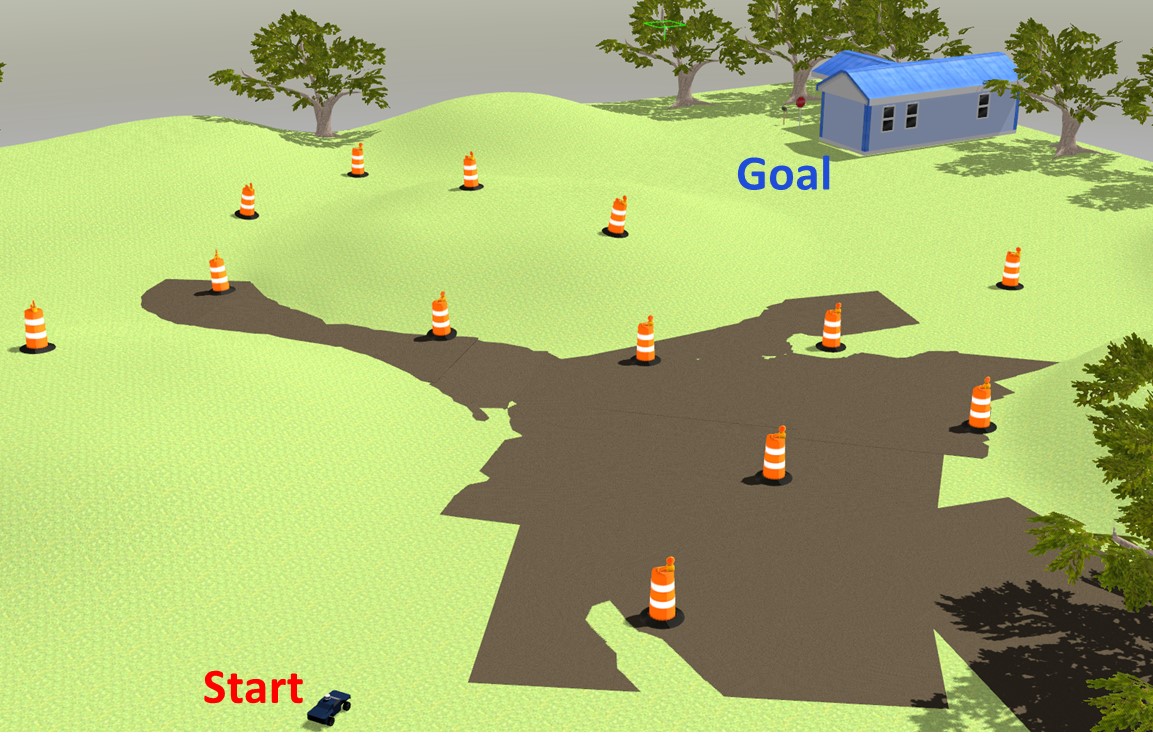}
\caption{Gazebo simulation environment with start and goal position. 
}\label{fig:simenv}
\end{figure}

\begin{figure}[ht]
\centering
\includegraphics[width=0.45\textwidth]{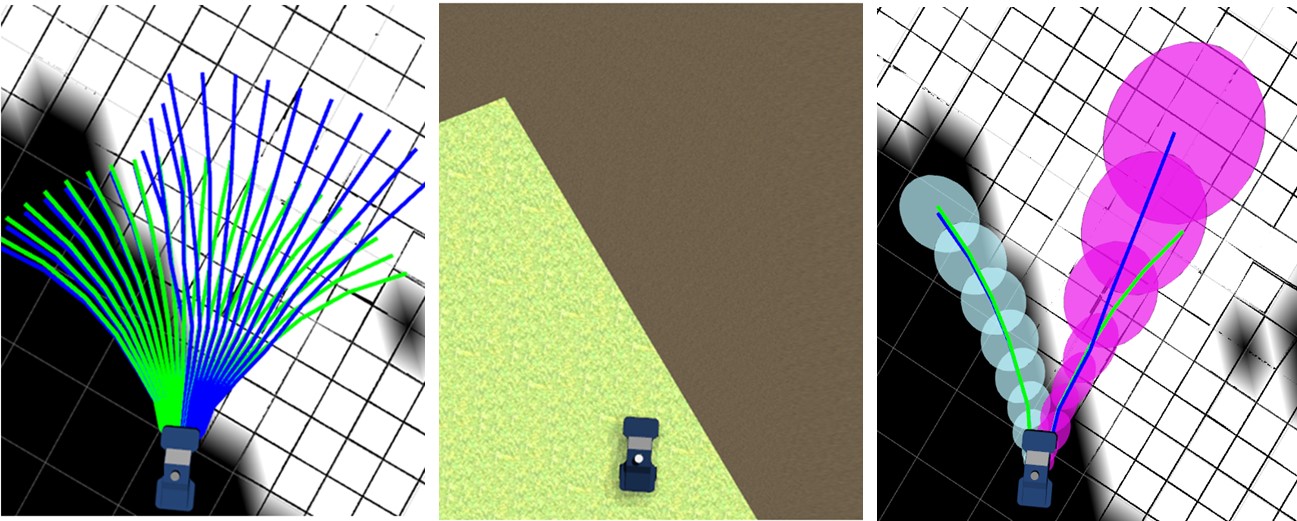}
\caption{Sampled nominal paths (green) and the mean of predictive path distributions (blue) are drawn on the left side. The predictive path distributions over the mud area (purple ellipsoid) and the grass area (cyan ellipsoid) are represented on the right side.
}\label{fig:pathsample}
\end{figure}

A modified version of the Gazebo environment in the AutoRally research platform is used \cite{goldfain2019autorally}. The vehicle is equipped with a front RGB camera and a 3D Lidar mounted on the vehicle. A 3D environment has hills and obstacles between the start and goal position, as illustrated in Fig. \ref{fig:simenv}.
There are two different types of terrains; i) the grass area, which has higher friction coefficient; and ii) the mud area, which has lower friction coefficient than the grass area. Correspondingly, we trained two different GP models for each terrain using GPyTorch \cite{gardner2018gpytorch}. The training data was collected with a joystick-controlled vehicle in the same environment while driving on the both grass and mud area. Due to the slip that occurs in the mud area, the deviation of the predictive path distribution from the nominal path is much higher in the mud area, as shown in Fig. \ref{fig:pathsample}. Therefore, we expect that much higher uncertainty costs, $e_{i}$, were assigned to the paths predicted over the mud area.

Instead of employing the 2D segmentation module like in \cite{leung2022hybrid}, we assumed each pixel were classified into two labels (i.e., grass and mud) depending on the color where the RGB color boundary was manually set. The elevation, terrain type, and geometric traversability maps have a grid size of $0.5m$ and an update rate of $10Hz$. The algorithm and the simulation is processed by a desktop computer with Intel i9-9900k CPU, 64GB RAM, and Nvidia RTX 2070 GPU. For uncertainty-aware path planning step, we set $I=63$, $M=20$, $R_{thres} = 0.3$, and the Gaussian kernel size as $s=5$. Also, the discrete system has a sample time of $0.1 sec$, and the prediction Horizon is $3 sec$ resulting in $N = 30$. Furthermore, during each MPPI path following process, we use sample time as $0.1 sec$, and the prediction horizon as $2 sec$. Local path was published at $3 Hz$ to the MPPI controller.

\subsection{Comparative Analysis}
To validate the performance of the proposed algorithm, we conduct a comparative analysis with two baseline navigation methods: i) a method that only considers the geometric traversability cost \cite{wermelinger2016navigation}; and ii) a method that considers the hybrid traversability cost, which accounts for both semantic and geometric traversability costs by simply adding two different costs \cite{leung2022hybrid}. For real-time path planning of the baseline navigation methods, we choose $A^{\ast}$ as the global path planner and Hybrid $A^{\ast}$ as the local path planner since it has been proven to be successfully applied for many off-road vehicle navigation \cite{thoresen2021path,guantns,dolgov2008practical,petereit2012application}. Pure pursuit controller is applied for path following of baseline navigation methods \cite{coulter1992implementation}. 
\begin{figure}[ht]
\centering
\includegraphics[width=0.49\textwidth]{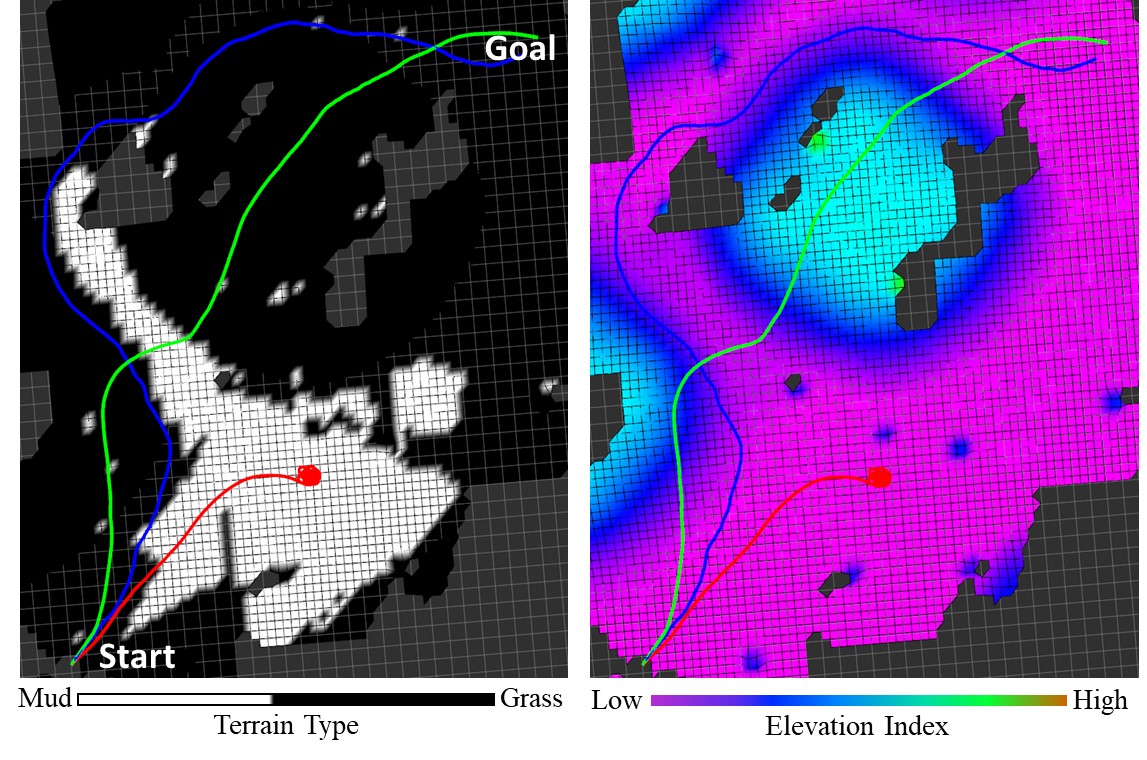}
\caption{Comparison of paths driven by the proposed algorithm (green), baseline1 method with geometric traversability cost only (red), baseline2 method with hybrid traversability cost (blue). Paths are illustrated on top of the processed terrain type map (left) and elevation map (right). Gray areas in both maps are undiscovered areas during the mission.
}\label{fig:simulationpaths}
\end{figure}

Each algorithm was tested 10 times with different locations of obstacles in the same environment setting of 3D terrain topology and terrain type. The simulation result of the example scenario is shown in Fig. \ref{fig:simulationpaths}. The first baseline navigation method generated the red path driving over the large mud area as it did not account for semantic features at all. The vehicle started to spin due to a high slip in the mud area and could not reach the goal in this run. The second baseline navigation method exhibited the blue path, which detoured the mud area and resulted in a too conservative longer path. Lastly, our method sometimes avoided or drove over the mud area to make safe but agile navigation. Remarkably, the proposed algorithm guided the vehicle to cross the mud area as it guarantees vehicle safety based on the predictive path distribution. This, in turn, results in a much shorter path than the others. 

The overall simulation statistics are shown in Table \ref{tb:simultion}. The qualities of paths are characterized by the average path length of successful runs and the number of successful runs, i.e., no collision and rollover. The path generated by the proposed method is shorter than the baselines while it successfully navigated to the goal point.

\begin{table}[htpb]
\caption{The overall simulation  results}
\label{tb:simultion}
\begin{center}
\begin{small}
\begin{sc}
\begin{tabular}{lcccr}
\toprule
Methods  & Avg.Path & Successful runs\\ 
   &Length        &(out of 10 trials)   \\
\midrule
Baseline1 (red)   & 61.2  & 4   \\
Baseline2 (blue)  & 65.7 & 8  \\
Proposed (green) & 52.6 & 10 \\
\bottomrule
\end{tabular}
\end{sc}
\end{small}
\end{center}
\end{table}

\section{CONCLUSIONS}
This work proposed a novel uncertainty-aware navigation algorithm to cope with 3D terrain topology and varying terrain conditions. Our approach leverages a learning-based framework that incorporates the terrain-induced uncertainty into the traversability cost and further preserves the realtimeness based on sampling-based methods. We validated the proposed algorithm in a 3D simulation environment with different types of terrains and showcased safe and agile navigation performance. In the future work, we will test our method on a physical vehicle in real-world off-road environments. For further improvement in the algorithmic aspect, we also plan to extend our work by incorporating an uncertainty-aware global path planner to avoid the local minima problem.

%\addtolength{\textheight}{-12cm}   % This command serves to balance the column lengths
                                  % on the last page of the document manually. It shortens
                                  % the textheight of the last page by a suitable amount.
                                  % This command does not take effect until the next page
                                  % so it should come on the page before the last. Make
                                  % sure that you do not shorten the textheight too much.

%%%%%%%%%%%%%%%%%%%%%%%%%%%%%%%%%%%%%%%%%%%%%%%%%%%%%%%%%%%%%%%%%%%%%%%%%%%%%%%%

%%%%%%%%%%%%%%%%%%%%%%%%%%%%%%%%%%%%%%%%%%%%%%%%%%%%%%%%%%%%%%%%%%%%%%%%%%%%%%%%

%%%%%%%%%%%%%%%%%%%%%%%%%%%%%%%%%%%%%%%%%%%%%%%%%%%%%%%%%%%%%%%%%%%%%%%%%%%%%%%%

\bibliographystyle{IEEEtran}
\bibliography{references}

\end{document}